\documentclass{article}

\usepackage[numbers, compress]{natbib}

    \usepackage[preprint]{neurips_2025}

\usepackage[utf8]{inputenc} 
\usepackage[T1]{fontenc}    
\usepackage{hyperref}       
\usepackage{url}            
\usepackage{booktabs}       
\usepackage{amsfonts}       
\usepackage{nicefrac}       
\usepackage{microtype}      
\usepackage{xcolor}         
\usepackage{multirow}
\usepackage{amsmath}
\usepackage{graphicx}
\usepackage{tabularx}
\usepackage{svg}

\title{REMOR: Automated Peer Review Generation with LLM Reasoning and Multi-Objective Reinforcement Learning}

\author{%
  Pawin Taechoyotin$^{1}$\quad
  Daniel Acuna$^{1,2,3}$\\
  \texttt{pawin.taechoyotin@colorado.edu}\quad
  \texttt{daniel.acuna@colorado.edu}\\[0.8ex]
  $^{1}$Department of Computer Science, University of Colorado Boulder\\
  $^{2}$Department of Information Science, University of Colorado Boulder\\
  $^{3}$ReviewerZero AI Inc., Boulder, CO
}

\begin{document}

\maketitle

\begin{abstract}
AI-based peer review systems tend to produce shallow and overpraising suggestions compared to human feedback. Here, we evaluate how well a reasoning LLM trained with multi-objective reinforcement learning (REMOR) can overcome these limitations. We start by designing a multi-aspect reward function that aligns with human evaluation of reviews. The aspects are related to the review itself (e.g., criticisms, novelty) and the relationship between the review and the manuscript (i.e., relevance). First, we perform supervised fine-tuning of DeepSeek-R1-Distill-Qwen-7B using LoRA on PeerRT, a new dataset of high-quality top AI conference reviews enriched with reasoning traces. We then apply Group Relative Policy Optimization (GRPO) to train two models: REMOR-H (with the human-aligned reward) and REMOR-U (with a uniform reward). Interestingly, the human-aligned reward penalizes aspects typically associated with strong reviews, leading REMOR-U to produce qualitatively more substantive feedback. Our results show that REMOR-U and REMOR-H achieve more than twice the average rewards of human reviews, non-reasoning state-of-the-art agentic multi-modal AI review systems, and general commercial LLM baselines. We found that while the best AI and human reviews are comparable in quality, REMOR avoids the long tail of low-quality human reviews. We discuss how reasoning is key to achieving these improvements and release the Human-aligned Peer Review Reward (HPRR) function, the Peer Review Reasoning-enriched Traces (PeerRT) dataset, and the REMOR models, which we believe can help spur progress in the area.
\end{abstract}

\section{Introduction}
Scientific peer review seeks to guide modern science by assessing manuscripts across multiple aspects, including novelty, methodological rigor, and potential impact \cite{kelly2014peer, morley2021now}. Typically, manuscripts are evaluated by multiple reviewers to capture diverse perspectives and provide authors with thorough feedback \cite{sani2023journal}. Such varied assessments can ideally offer authors fresh insights into their work, helping them address potential weaknesses. Ultimately, peer review aspires to identify research that genuinely matters to the broader scientific community, though achieving this goal consistently remains challenging.

There are multiple problems with the scientific peer review that could be improved. The process tends to linger too long, the reviewers have biases against certain topics, and other scientists \cite{kadaifci2025fundamental, resnik2008perceptions}. Other problems include the lack of incentive to review, low review panel diversity, and small expert availability \cite{el2023critical, shah2022challenges}. With the exponential increase in submissions across science, these issues are exacerbated, creating "reviewer fatigue" for experts \cite{breuning2015reviewer}. Less-experienced reviewers end up writing reports with unhelpful feedback \cite{karlberg2015peer, d2017can}. The scientific community needs to experiment with new ways of scaling high-quality scientific peer review.

In recent years, Large Language Models (LLMs) have demonstrated remarkable capabilities in understanding and generating text. The models perform well in text summarization, question answering, coding, and more \cite{shao2024survey, zhang2024benchmarking}. It is only natural that these capabilities can translate into automating peer review. Recent work has shown how useful LLMs can be for this task (e.g., \cite{liang2024can, d2024marg, taechoyotin2024mamorx, zhuang2025large}). The advances have been encouraging, but similar failures keep appearing across models and systems, calling for improvements.

\paragraph{Issues with current AI review generation systems} The first issue concerns the depth of the review. It has been found that the generated comments overpraise the work, are too generic, and overlook details \cite{yuan2022can, checco2021ai}. The second issue involves formalizing and operationalizing what a good peer review should be. Currently, there have been many studies on the components of an ideal peer review \cite{Wang2023, Superchi2019, 10.1371/journal.pbio.3002238, Superchie035604}, but the metrics have not been operationalized for AI. On the other hand, many frameworks have been proposed for evaluating a manuscript and producing the peer review \cite{kirtani2025revieweval, d2024marg, zhou2024llm}, but have the limitations of requiring human oversight. Therefore, the issues revolve around depth, operationalized human-aligned evaluation, and effective automation.


\begin{figure}[htbp]
  \centering
  \includegraphics[width=1\textwidth]{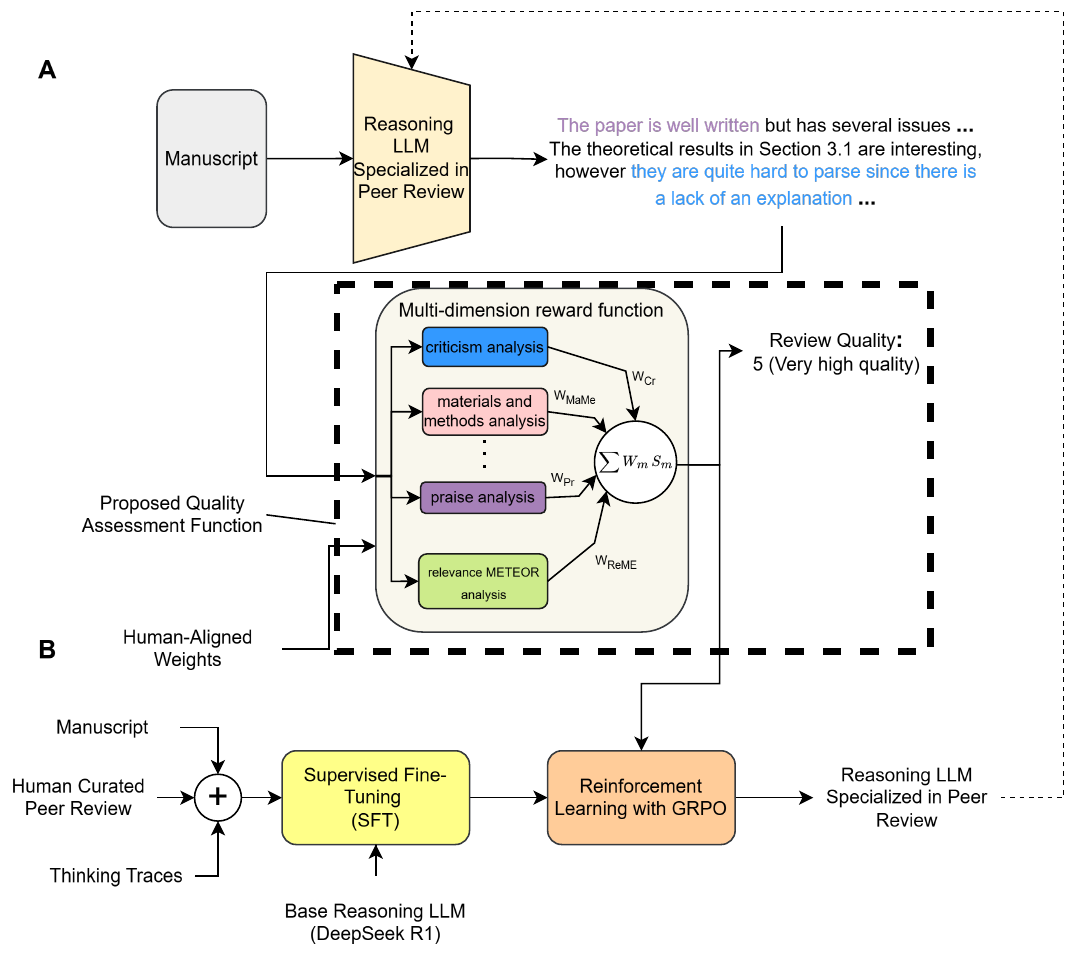}
  \caption{\textbf{Illustrative diagram of how REMOR works.} \textbf{A.} AI-review generation system with our proposed multi-dimensional reward function (HPRR), which evaluates the quality of the peer review. A sample review is shown and color-coded to indicate the relevance between the analysis metric and the review. For this instance, purple indicates content related to praise, and blue is content related to criticism.  \textbf{B.} The training is performed on a dataset (PeerRT) comprising the manuscripts, human peer review, and synthetic thinking traces. The base model is first trained via supervised fine-tuning (SFT) and later by GRPO with a multi-dimensional reward function. Finally, we use human-aligned weights to create a linear combination of the multi-dimensional reward for GRPO.}
  \label{fig:automated-peer-review-generation-system}
\end{figure}

In this work, we propose REMOR (AI-Generated Peer Reviews through Reasoning and Multi-Objective Reinforcement Learning). Figure \ref{fig:automated-peer-review-generation-system} shows a diagram outlining our work. REMOR is a framework that uses reasoning and a multi-objective reward strategy. The reasoning is meant to increase the depth of the review. The multi-objective part is meant to operationalize what a good review should look like. We begin with a model fine-tuned on a novel dataset we call the Peer Review Reasoning-enriched Traces (PeerRT). The fine-tuning step enables the model to learn domain-specific language styles and common review structures. We then perform Reinforcement Learning (RL) with GRPO using a new reward function aligned to human preferences across multiple aspects of review quality, including detailed criticisms, tangible examples, importance-based prioritization, and well-grounded suggestions. We call this function Human-Aligned Peer Review Reward (HPRR). We develop two models, one that uses human-aligned human preferences (REMOR-H) and another that is uniform (REMOR-U). We quantitatively and qualitatively show that the reviews generated by REMOR achieve twice as high rewards as those generated by humans and other agentic and non-agentic systems. In sum, our work provides the following contributions:
\begin{itemize}
    \small
    \item PeerRT: An enriched dataset of peer reviews with reasoning traces and metrics in each aspect.
    \item Human-aligned Peer Review Reward (HPRR): An operationalized metric for RL that quantifies the quality of peer review feedback.    
    \item REMOR: An LLM reasoning model that specializes in peer review.
    \item A comparison of multiple non-reasoning, agentic, and general reasoning models against REMOR-U and REMOR-H
    \item A qualitative analysis of reviews generated by REMOR.
\end{itemize}

\section{Related Work}

\subsection{Non-reasoning AI-based Peer Review Generation Systems}
Automated scientific review generation has seen significant advancements in recent years thanks to LLMs \cite{yuan2022can, bartoli2016your, wang2020reviewrobot, liu2023reviewergpt, d2024marg, taechoyotin2024mamorx}. It has been shown that LLMs are highly capable but contain limitations such as a lack of personalization, context-awareness, and subtle critiques often found in human feedback \cite{liang2024can}. Potential biases also exist within the models, such as overpraising the work and providing shallow suggestions. There have been promising efforts in mitigating these limitations, such as engineering system prompts to add personalization and structure to the review \cite{liang2024can}. Currently, the most promising systems are based on a panel of AI agents. In \cite{d2024marg}, the authors created sets of agents that focus on different parts of the review, such as experiment quality, overall clarity, and work impact, with later sets of agents refining the feedback. In the recent work of \cite{taechoyotin2024mamorx}, the authors used multiple agents with large context windows, each attending to different aspects of the article, such as novelty, impact, and figures (the model was multi-modal), with an "area chair" agent that condensed the suggestions. The authors show that the large context windows of current models, their multi-modality, and their ability to call external tools significantly improved review quality.

One downside of AI agent panels is that they take excessively long to coordinate and exchange contextual information. In the recent work of \cite{taechoyotin2024mamorx, d2024marg}, the agents spend 20 minutes or more creating the review, using millions of tokens, many of them solely for coordinating. In contrast, non-agent systems take significantly less (e.g., less than 1 minute in \cite{liang2024can}). Still, agentic systems are perceived as significantly better than human reviews and non-agentic systems \cite{taechoyotin2024mamorx}.

\subsection{Evaluating peer review generation systems}

Evaluating the quality of generated peer reviews is complex and subjective. In \cite{d2024marg}, the authors used a survey asking humans to determine whether the comments were relevant. In \cite{taechoyotin2024mamorx}, the authors created an arena-style match system to ask humans to choose between pairs of AI-generated reviews. The same quality evaluation issues that plague human reviews also affect AI: it is challenging to develop guidelines.

Other work has sought to automate parts of the evaluation system. Similar to how LLMs are evaluated for other tasks, scientists have used translation quality metrics such as ROUGE \cite{lin2004rouge}, BERTScore \cite{zhang2019bertscore}, and METEOR \cite{banerjee2005meteor} to assess how closely the reviews discuss topics present in the manuscript or align with human-generated reviews. Automated evaluations can also use LLMs themselves as LLM-as-judge \cite{zheng2023judging} or even use an LLM to address the applicability and relevance of suggestions \cite{kirtani2025revieweval, d2024marg}. A more systematic approach to evaluating AI-generated peer review was proposed by \cite{kirtani2025revieweval}. In such a study, the authors suggest an evaluation that focuses on evaluating comparisons with human reviews, factual accuracy, analytical depth, and actionable insights. The goal is to detect superficial critiques, hallucinations, and next-step feedback.

The overall issue with evaluation is that most methods assume a human review is available. As AI systems improve, good-quality human reviews to continue guiding AI are scarce. One possibility is to take principles of a good review and create an automated evaluation based on them. If we use such an automated method, we could drive a system to achieve "superhuman" performance, optimizing a review along all the aspects of what is considered a desirable review.

\subsection{Reasoning, Reinforcement Learning, and GRPO}

Chain-of-Thought Prompting showed a remarkable ability to improve the performance of existing systems \cite{NEURIPS2022_9d560961}. Through this experience, the idea of building "chains of thought" directly in the generation resulted in "reasoning" models. Reasoning systems have become increasingly dominant in LLM leaderboards (e.g., top places in llmarena.ai \cite{chiang2024chatbot}). One of the first applications of this idea was in GPT-3 where human feedback was used to train a reward model that reinforcement learning can use. This technique is called Reinforcement Learning with Human Feedback (RLHF) \cite{christiano2017deep}. RLHF can be used to generate models that go beyond next-token generation. For example, it can generate instruction-following GPTs \cite{NEURIPS2022_b1efde53}. 

Early reasoning ideas relied on explicit differentiable reward models \cite{schulman2017proximal, silver2017mastering}, such as Proximal Policy Optimization (PPO). In other words, the reward model needed to be a neural network used appropriately by reinforcement learning. However, these models are hard to define and optimize. In \cite{shao2024deepseekmath}, the authors proposed a novel method for reinforcement learning that did not require an explicit model. The method proved to improve memory usage and performance of training LLMs with reinforcement learning. The method is called Group Relative Policy Optimization (GRPO). Importantly, this technique only needs the reward values of generation traces to optimize the underlying policy, which is much more flexible.

\section{Reasoning and Reinforcement Learning for Automated Peer Review Generation}

Our current work uses recent developments in reasoning LLMs. In particular, we use supervised fine-tuning on high-quality reviews (in the spirit of \cite{muennighoff2025s1}), followed by a multi-objective reinforcement step that tries to mimic what humans like in a review. We now review the datasets used and later describe the method.

\subsection{Peer review datasets and models}

\subsubsection{Data curation}
We base our dataset on readily available review data from ICLR 2017-2020 \cite{Wang2023, kang2018dataset}. The dataset contains 16.8K reviews across 5.5K submissions. We augmented the data with the full text of each paper in OpenReview. The full text is extracted from a PDF file using GROBID \cite{lopez2009grobid, romary2015grobid}. Additionally, we augmented the reviews with reasoning traces generated from Claude Sonnet 3.7 extended thinking model. The thinking trace is meant to force the supervised fine-tuned (SFT) model to generate thinking traces. We call this final dataset the Peer Review Reasoning-enriched Traces (PeerRT). The prompt templates are in the Appendix \ref{app:prompt-templates}.

\subsubsection{Models}
We use the Claude Sonnet 3.7 extended thinking model as a base, large and proprietary LLM model, and DeepSeek-R1 (7B distillation) \cite{guo2025deepseek} model to compare against. We further use the DeepSeek-R1 model during the supervised-fine-tuning (SFT) and reinforcement learning via group relative policy optimization (GRPO) phases.

\subsubsection{Reward Model}
\label{sec:reward_model}
Our reward model combines a sentence-level review assessment model and a relevance score between the generated review and the reviewed manuscript. The sentence-level review assessment is based on the work of \cite{10.1371/journal.pbio.3002238}. It evaluates a sentence along several aspects of what a review should contain: criticism, example, importance and relevance, materials and methods, presentation and reporting, results and discussion, suggestion and solution. We normalize each aspect by the number of sentences in the review. To measure the review's relevance to the manuscript, we use the METEOR score between the review and the manuscript's text \cite{banerjee2005meteor}. The final reward is the weighted sum of the eight normalized quality metrics and METEOR.

\subsection{Multi-objective reinforcement learning}

Because our reward model contains multiple aspects, we must combine them to make them usable within GRPO. There are multiple methods for achieving this combination. In this article, we use a weighted sum of the rewards to achieve a multi-objective reinforcement learning (MORL) that is Pareto-optimal \cite{castelletti2013multiobjective, li2020deep, zhang2024novel}. The first weighted structure is Uniform (U). The second weighted structure is called Human-aligned weights (H), where the values are computed from the human preferences in \cite{taechoyotin2024mamorx}. The basic idea is to boost aspects that humans prefer when comparing reviews (see below for details).

\subsubsection{Obtaining reward weights from human preferences}
\label{sec:obtaining-rewards-weights-from-human-value}

Here we detail how to find the weights of the human-aligned reward. We use the dataset of human preferences described in \cite{taechoyotin2024mamorx}, which results from an arena-style competition across multiple human and AI-generated reviews. In particular, for a paper $p$ and a reviewer $k$, we have pairs of reviews $\text{review}_a$ and $\text{review}_b$ from two systems and the user vote of whether "A is better than B", "B is better than A", or "A is equal to B". In this task, "better" was defined as having "overall" higher quality. We compute the set of dimensions of the reward system based on \ref{sec:reward_model} as covariates and use a Bradley-Terry (BT) Model \cite{19ff28b9-64f9-3656-ba40-08326a05748e} and two other variations to understand how to weigh each review aspect to maximize the predictive accuracy. We now explain these models:

\paragraph{Adapted Bradley-Terry Model (ABT)} The adapted Bradley-Terry Model explicitly models ties as a three-outcome regression. We further constrained the weights to be positive. (In the standard BT analysis, we found that some weights were negative, which is not qualitatively sensible \cite{19ff28b9-64f9-3656-ba40-08326a05748e})

\paragraph{Constrained Reward Model (CRM)} This model is similar to ABT, where an L1 regularization on the weights is applied to avoid overfitting (see Appendix \ref{app:crm-definition}).

After obtaining the weights from each algorithm, we scale them with a min-max scaler and reweigh them to add up to nine. To avoid an aspect being discarded in our Multi-objective Reinforcement Learning, we apply Laplace smoothing with $\alpha = 0.01$. The unadjusted weights can be found in Appendix \ref{app:human-value-weight-unadjusted}.

\section{Experimental results}

\subsection{Experimental Setup}
\label{sub-sec:experimental-setup}
All training was performed on a single virtual server with 64 vCPUs, 256 GB of RAM, and 2 NVIDIA A100 80GB GPUs. Our preliminary results show that training on high-quality samples results in the best performance. The selection is done by computing each sample's uniform reward (\ref{sec:reward_model}) and selecting the samples with a score above the 90th percentile.  The selected samples are used for supervised fine-tuning (SFT) and reinforcement learning via Group Relative Policy Optimization (GRPO). The code and dataset links are in Appendix \ref{app:code-repository} and \ref{app:trained-models-and-dataset}.

\subsubsection{Software setup}
We used DeepSpeed \cite{rajbhandari2020zero} with LLaMA-Factory \cite{zheng2024llamafactory} and TRL to train our models. DeepSpeed enables splitting optimizer states of the model across multiple heterogeneous devices. This enabled the training of LLMs to a broader population with limited computing resources. Essentially, it enables training of LLMs that do not fit within a single GPU to be trained across multiple GPUs. In most cases, it also enables scientists to train LLMs using multiple smaller GPUs instead of just a single large GPU. Supervised fine-tuning may outperform test-timing reasoning, given enough computational resources \cite{snell2024scalingllmtesttimecompute}. 

\subsubsection{Supervised fine-tuning}
To simplify supervised fine-tuning (SFT), we use LLaMA-Factory \cite{zheng2024llamafactory}. LLaMA-Factory is a unified framework that provides tools and implementation of techniques to finetune LLMs efficiently. We trained DeepSeek-R1-Qwen-Distilled-7B via LoRA with rank = 8, cutoff length = 32,768, gradient accumulation steps = 4, learning rate = 0.0001 for three epochs. The training time on 1.7k samples is approximately 6.5 hours.

\subsubsection{Multi-objective RL + GRPO}
Due to LLaMA-Factory not supporting GRPO, we used the TRL software package to perform multi-objective reinforcement learning with GRPO on our LLMs. The parameters used for training are generations = 4, max prompt length = 12,288, max completion length = 4096, device batch size = 4, gradient accumulation steps = 1, and temperature = 0.9 for one epoch. The other non-specified parameters are kept at their default value. The training time on 1.7k samples was approximately 100 hours for one epoch. The 1.7k samples and the aforementioned parameters resulted in 864 total optimization steps.

\subsection{Human-aligned reward vs. uniform rewards}
We estimate the best weights with various algorithms as described in section \ref{sec:obtaining-rewards-weights-from-human-value}. The Adapted Bradly-Terry obtained the best five-fold cross-validation performance for predicting human preferences (Appendix Table \ref{tab:human-value-weights-laplace-smoothing}). The results indicate that humans mostly favor "Importance and Relevance", "Suggestions and Solution", and "Relevance" (METEOR) with a weight of 0.11, 0.16, and 8.67, respectively. Other metrics were irrelevant, with a weight of 0.01 (the Laplace smoothing parameter). The final human-aligned weights can be found in Appendix \ref{app:human-value-weight-unadjusted}, Table \ref{tab:human-value-weights-laplace-smoothing}.

\subsection{Reasoning model results}
\begin{figure}[htbp]
  \centering
  \includegraphics[width=1\textwidth]{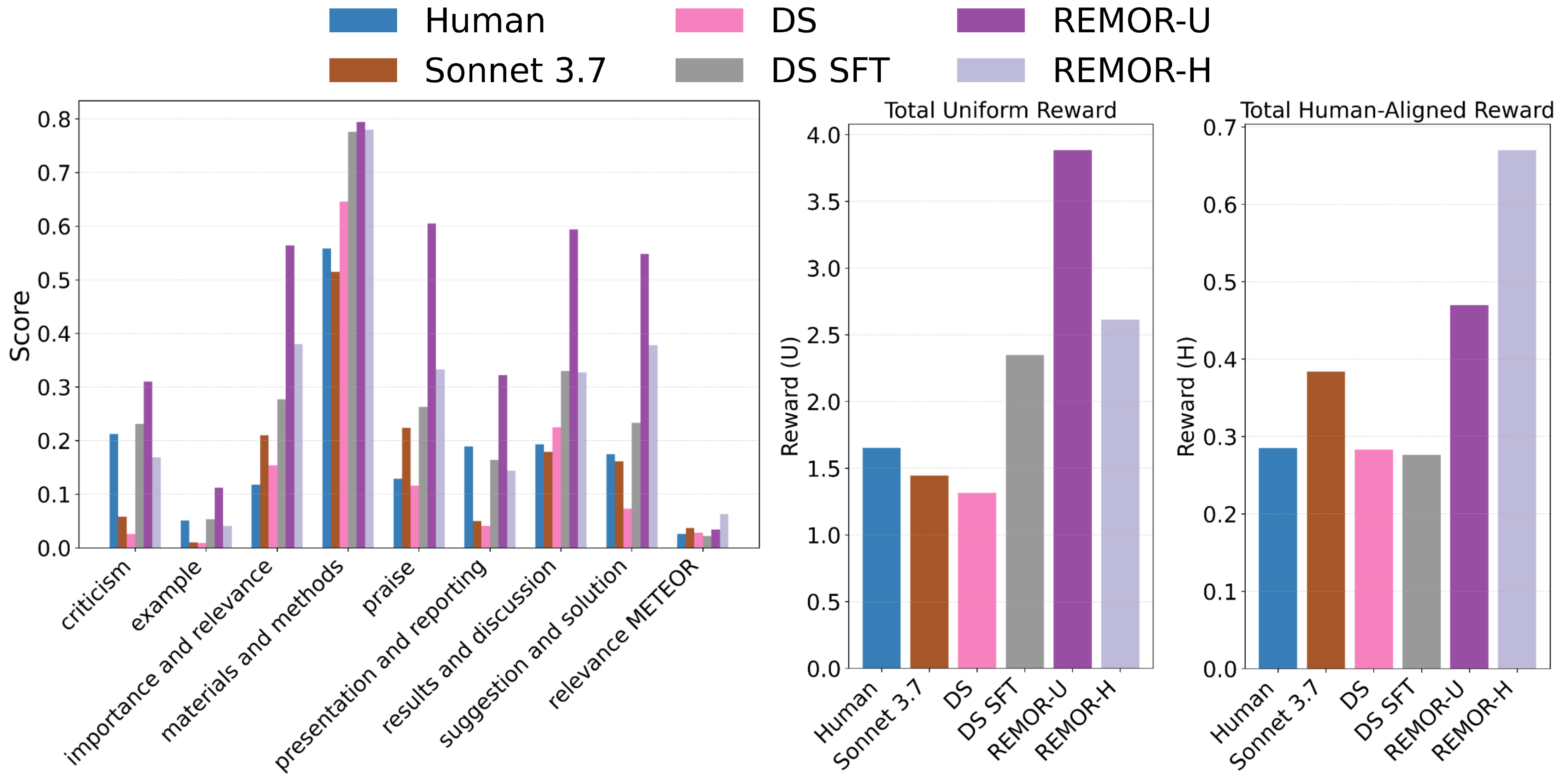}
  \caption{The left panel shows the average reward across the metrics for each model. We can see that REMOR-U has the top reward across all metrics and has the second highest score in METEOR. The middle and right panels show each model's total uniform and human-aligned reward. REMOR-U significantly outperforms all models in total uniform reward. For total human-aligned reward, REMOR-H has the highest score. All values had a SEM within the range of 0.001. (See Table \ref{tab:metric-wise-comparison-of-training-treatments} for full results in the Appendix.)}
  \label{fig:reasoning-model-results-bar-chart}
\end{figure}

The performance of all models is shown in Figure \ref{fig:reasoning-model-results-bar-chart}. We have included the average score across different metrics for each model and the uniform and human-aligned weighted reward. The pretrained model DeepSeek-R1 (DS) has the lowest total rewards at 1.317 and 0.283 for uniform reward and human-aligned reward, respectively. With supervised fine-tuning, the DeepSeek-R1 model improved dramatically and outperformed both Sonnet 3.7 and human reviewers in uniform reward but slightly underperformed with human-aligned reward. After applying reinforcement learning with GRPO to the SFT model, the performance drastically improved once more. In this instance, both GRPO model variants outperformed humans and Sonnet 3.7 by a significant margin. The model DeepSeek-R1, which was trained with the uniform reward, is called REMOR-U. This model had the highest uniform reward at 3.884. The model DeepSeek-R1, which was trained with the human-aligned reward, is called REMOR-H. For obvious reasons, REMOR-H loses to REMOR-U regarding uniform reward but has the highest human-aligned reward at 0.670. A variance analysis across metrics shows that REMOR-U can excel at multiple metrics simultaneously, while other models only focus on one metric at a time (Figure \ref{fig:reasoning-model-results-bar-chart}). A distribution analysis of the uniform reward shows that on average REMOR-U has a higher score than human reviews (see Appendix Figure \ref{fig:reward-distribution-human-vs-remor}).

\subsection{Learning Curve For GRPO}
\begin{figure}[htbp]
  \centering
  \includegraphics[width=1\textwidth]{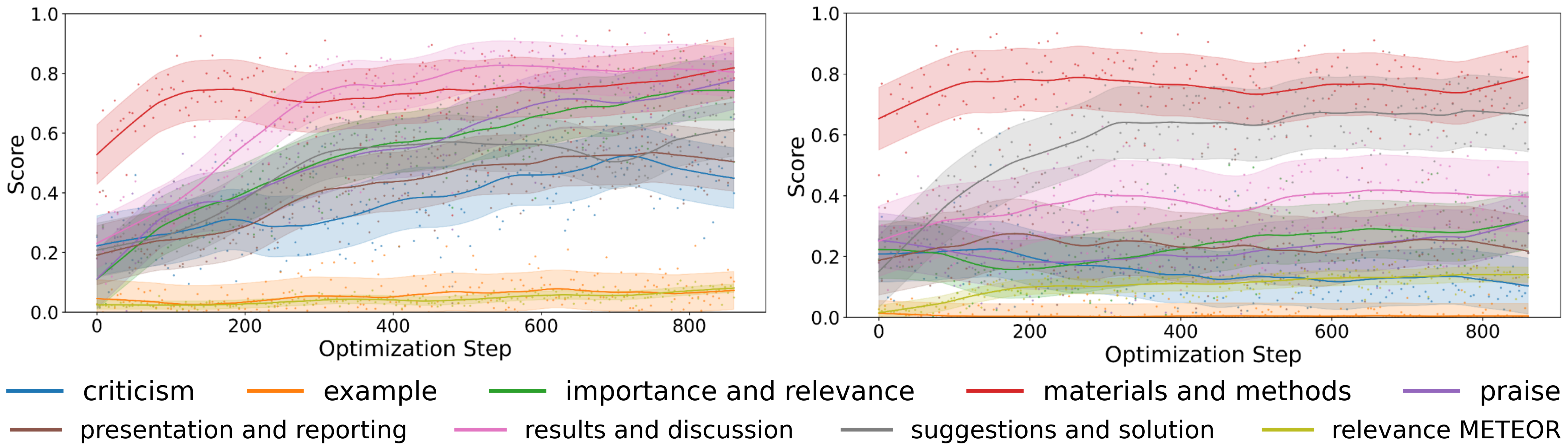}
  \caption{Reward for each aspect per each optimization step. The left panel depicts the reward trend for all metrics of REMOR-U; the right panel depicts the reward trend for all metrics of REMOR-H. The vertical axis depicts the reward value from 0 to 1. The horizontal axis depicts the optimization step.}
  \label{fig:grpo-metrics-progression}
\end{figure}

Figure \ref{fig:grpo-metrics-progression} shows the average reward across each metric. The left and right panel show the improvement for REMOR-U and REMOR-H, respectively. The reward improvement trend of REMOR-U gives us insight into the difficulty of improving each metric. We can divide the metrics into groups: hard, moderate, and easy. The hard difficulty has a peak reward at 0.1. The moderate difficulty has a peak reward within the range of 0.4 to 0.6. The easy difficulty has a peak at 0.8. This insight can lead to a normalization scheme for each metric in future studies that factor in the different nature of each metric. The relevance METEOR metric is an example where a score of 1.0 is not ideal since that requires the review to be excessively long to contain all the contents within the manuscript. The relevance METEOR is in the hard difficulty group.


\subsection{Non-reasoning, agentic systems for peer review generation}
\begin{figure}[htbp]
  \centering
  \includegraphics[width=1\textwidth]{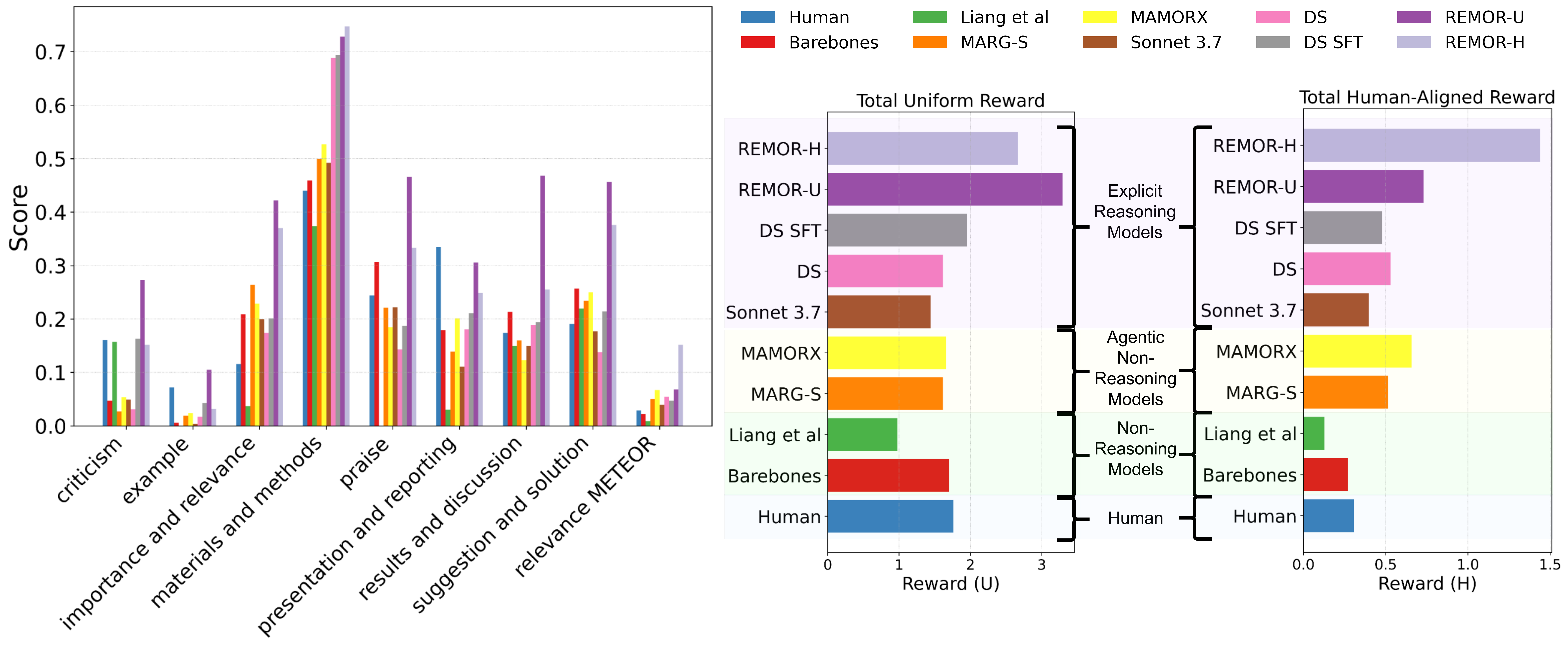}
  \caption{The comparison shows that REMOR outperforms non-reasoning models across all metrics.
  The left panel shows that REMOR-U has the highest scores across metrics compared to other models. The middle and the right panel show that REMOR outperforms all models in uniform reward and human-aligned reward.
  All values had a SEM within the range of 0.001. (Full results in Appendix Table \ref{tab:metric-comparison-of-previous-systems})}
  \label{fig:performance-comparison-against-agentic-systems}
\end{figure}

We benchmark REMOR with other peer-review generation systems using papers from various conferences. The results are shown in Figure \ref{fig:performance-comparison-against-agentic-systems}. Both variants of REMOR significantly outperformed human reviewers, all non-reasoning model systems, and other reasoning models in terms of uniform and human-aligned reward. REMOR-U had the highest average uniform reward at 3.292. REMOR-H has the highest average human-aligned reward at 1.438. REMOR-U also has the highest average score in "criticism", "example", "importance and relevance", "praise", "presentation and reporting", "result and discussion", "suggestion and solution" with a score of 0.273, 0.105, 0.422, 0.466, 0.306, 0.468, and 0.456 respectively. Interestingly, "materials and methods" of REMOR-H has a higher score than "importance and relevance," even though "importance and relevance" has around 10 times higher weight value. Additionally, the results in Figure \ref{fig:performance-comparison-against-agentic-systems} indicate that reasoning models outperform non-reasoning agentic AI systems such as MARG-S \cite{d2024marg} and MAMORX \cite{taechoyotin2024mamorx}.

\subsection{Variance analysis}

\begin{figure}[htbp]
  \centering
  \includegraphics[width=0.8\textwidth]{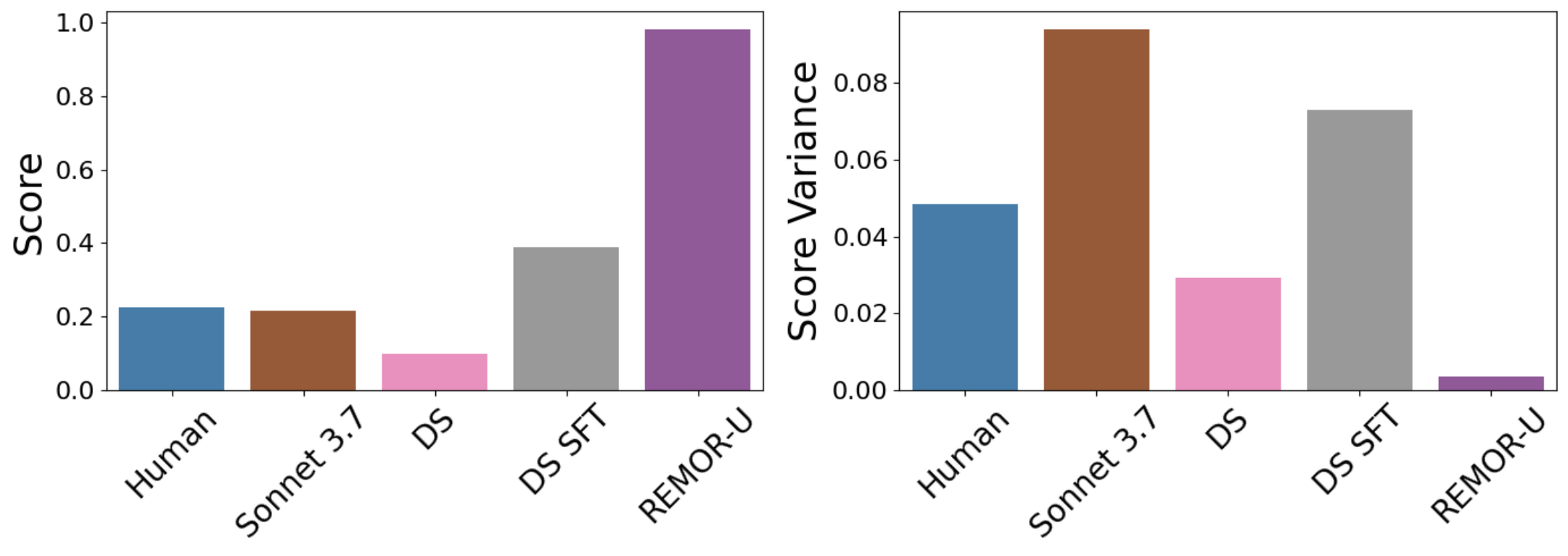}
  \caption{The left panel shows the normalized mean across metrics. We can see the REMOR-U significantly outperforms other systems, while the fine-tuned model (DS SFT) is the second best, with half the score of REMOR-U. The right panel shows the variance across the metrics. The chart shows that REMOR-U has very low variance. Overall, REMOR-U has the highest mean and lowest variance across metrics, which indicates it is the best-performing model.}
  \label{fig:normalized-mean-analysis}
\end{figure}

We wanted to understand how robustly REMOR-U, the best model, produced the reward. Maybe it produced high reward with high variance, making it unreliable. We performed a normalized mean and variance analysis and found that REMOR has both the highest mean and lowest variance in rewards. This should lead to more consistent review outcomes (Figure \ref{fig:normalized-mean-analysis}).

\subsection{Qualitative Analysis}
We qualitatively analyze the reviews to assess differences between our REMOR models. We found that the highest-reward human and REMOR-U reviews contain high-quality comments that contribute to the improvement of the paper (e.g., "The paper is well written but...", "There is a lack of explanation in section 3.1 ...". Sample reviews can be found in (Appendix Table \ref{tab:sample_reviews_with_high_or_low_scores}). The comments provided by REMOR-U are direct and concise. Figure \ref{fig:performance-comparison-against-agentic-systems} shows that REMOR-U and REMOR-H should be similar quantitatively. However, we found that REMOR-H reviews tend to be too long and redundant. We believe this is caused by an excessive focus on the METEOR reward component, which attempts to align the review produced with the manuscript. Even though the comments are similar to REMOR-U, the review as a whole is difficult to understand. Therefore, we judge REMOR-U as qualitatively better than REMOR-H. Overall, we found that a higher uniform reward correlates with higher-quality reviews. Future studies should confirm this with a human comparison to reinforce this claim.

\section{Discussion}
In this article, we explore how reasoning models can help improve AI-generated reviews. We develop a multi-aspect reward function that feeds into Reinforcement Learning with GRPO. Once fine-tuning and RL GRPO are in place, the performance of the model in regards to multi-aspect reward function increased dramatically, showing that RL GRPO can surpass human performance for our tasks. 

We found that human preferences do not favor criticism despite multiple guidelines encouraging this type of feedback \cite{kelly2014peer}. We believe this is because criticism is perceived as a net negative \cite{nobarany2015use}. In contrast, humans prefer the "importance and relevance" aspect because they consider it crucial that novelty and impact assessment are part of a review \cite{teplitskiy2022novel, gallo2018risk}. A similar pattern is found for the "suggestion and solution" aspect. Lastly, we think "relevance" (METEOR) is highly favored because the comments directly relate to the work. Also, relevancy is a standard metric found in previous studies regarding peer review \cite{wilcox2019rude, liang2024can, d2024marg}. Thus, there can be significant differences between human preferences and the principles of a good peer review.

We found that general reasoning models (non-finetuned) performed comparably to multi-agentic review generation systems. However, with the addition of RL GRPO, the specialized reasoning models substantially outperform complex agentic systems like MARG-S \cite{d2024marg} and MAMORX \cite{taechoyotin2024mamorx}. A byproduct of this new architecture comes in processing time. Agentic systems are notoriously slow (approximately 20 - 30 minutes \cite{taechoyotin2024mamorx}) compared to REMOR (approximately 1 minute). Thus, a reasoning model with simple prompts is a promising alternative to augmenting the peer review process with LLMs.

Overall, our results show that reasoning in automated review generation systems leads to excellent performance improvements. We found that our training scheme effectively used the reasoning traces that allowed it to optimize multiple metrics simultaneously. Our system achieves better performance than other systems that use even more information, such as MAMORX \cite{taechoyotin2024mamorx}, which uses figures and external knowledge. Notably, while other systems appear to be doing reasoning in the form of chain-of-thought and multi-agent coordination (e.g., \cite{d2024marg}), it is not the type of modern reasoning used today. We also show how synthetic rewards can be effective for GRPO and training these systems.

\section{Limitations and Future work}
There are limitations in our study. Our set of reviews is narrow, and we are not using other modalities such as citations or figures. Previous studies have connected to external knowledge and multi-modality \cite{taechoyotin2024mamorx}, but we did not include this information. Furthermore, we could only support a limited context window, potentially leading to incomplete manuscript analysis or the impossibility of including raw external knowledge within the prompt. Also, our human-aligned reward model is based on a relatively small sample of votes. Finally, our evaluation is not based on direct human feedback but on a constructed human-aligned reward function. Therefore, a true blinded human evaluation by experts should confirm our findings. We believe these shortcomings are addressable, which we will do in future work.

\section{Conclusion}
In this work, we demonstrated that explicit reasoning combined with multi-objective reinforcement learning significantly enhances the depth and quality of automated peer review generation. By incorporating detailed, aspect-based reward signals, REMOR produced feedback that achieves approximately double the Human-aligned Peer Review Reward (HPRR) of typical human reviews, approaching parity with the best-quality human-generated reviews. Our proposed human-aligned reward function can also serve as a self-assessment tool for reviewers, encouraging more detailed and relevant feedback. To facilitate continued research, we release the PeerRT dataset enriched with reasoning traces, the HPRR evaluation metric, and our trained REMOR models. Future work will extend REMOR to incorporate multimodal information, such as figures and external references, and evaluate its applicability across broader scientific disciplines.

\bibliographystyle{plainnat}
\bibliography{references}

\newpage
\appendix

\section{Metric names}
Throughout the Appendix, we will use the following short names for each metric: criticism (Cr), example (Ex), importance and relevance (ImRe), materials and methods (MaMe), praise (Pr), presentation and reporting (PrRe), results and discussion (ReDi), and suggestion and solution (SuSo). The "relevance METEOR" score short name is ReME. The short names are meant to improve the ease of displaying results in tables.

\section{Human values weights calculation}
\label{app:human-value-weight-unadjusted}
This section includes the details for calculating the human-aligned weights and intermediate weights that have not been adjusted.

\subsection{Constrained Reward Model (CRM)} \label{app:crm-definition}
The Constrained Reward Model (CRM) is formulated as a constrained problem where convex optimization techniques are applied to find the human-aligned weights. We define $c$ as the weights, $d$ as the number of metrics (e.g. Cr, Ex, ImRe, etc.) , $s_x$ as the reward score for review $x$:

Let \( c \in \mathbb{R}^d \), where \( d = \text{number\_of\_metrics} \). 

\[
\begin{aligned}
\text{minimize} \quad & \sum_{i=1}^{d} c_i \\
\text{subject to} \quad 
& c_i \geq 0 \quad \forall i \in \{1, \dots, d\} \\
& \sum_{i=1}^{d} c_i \leq 1 \\
& \sum_{i=1}^{d} c_i \geq 0 \\
& \text{For each match } (a, b) \text{ in dataset:} \\
& s_a = \sum_{i=1}^{d} c_i a_i,\quad s_b = \sum_{i=1}^{d} c_i b_i \\
& \quad \text{where } a_i \text{ and } b_i \text{ are the covariates of reviews A and B, respectively} \\
& \text{Constraints based on match outcome:} \\
& \text{If the match is labeled as "A is better than B" } \text{ : } s_a \geq s_b + \varepsilon \\
& \text{If the match is labeled as "B is better than A" } \text{ : } s_a \leq s_b + \varepsilon \\
& \text{If the match is labeled as "A is equal to B" } \text{ : } s_a = s_b
\end{aligned}
\]

where \( \varepsilon = 10^{-12} \) is a small positive constant to ensure numerical stability for strict inequalities.

\begin{table}[htbp]
  \caption{Unadjusted human-aligned weights from each algorithm. F1 is the average F1 score computed from performing cross-validation.}
  \label{human-value-weights}
  \centering
  \small
  \begin{tabular}{ l | l | l | l | l | l | l | l | l | l || l ||}
    \toprule
    \textbf{Model} & \textbf{Cr} & \textbf{Ex} & \textbf{ImRe} & \textbf{MaMe} & \textbf{Pr} & \textbf{PrRe} & \textbf{ReDi} & \textbf{SuSo} & \textbf{ReME} & \textbf{F1}\\
    \bottomrule
    \multicolumn{11}{l}{\textbf{Bradley-Terry}}                  \\
    \bottomrule
    Logistic Regression & -0.20 & -0.74 & -0.14 & 0.22 & -0.11 & 0.05 & -0.05 & 0.54 & 1.21 & 0.43\\ 
    \bottomrule
    \multirow{2}{*}{\begin{tabular}[c]{@{}l@{}}Logistic Regression\\ Cross Validation\end{tabular}} & \multirow{2}{*}{-0.04} & \multirow{2}{*}{0.00} & \multirow{2}{*}{0.02} & \multirow{2}{*}{0.00} & \multirow{2}{*}{0.00} & \multirow{2}{*}{0.00} & \multirow{2}{*}{0.00} & \multirow{2}{*}{0.00} & \multirow{2}{*}{0.22} & \multirow{2}{*}{0.40}\\ 
    & & & & & & & & & &\\
    \bottomrule
    \multicolumn{11}{l}{\textbf{Adapted Bradley-Terry}}                   \\
    \bottomrule
    Linear Regression & 0.00 & 0.00 & 0.23 & 0.00 & 0.00 & 0.00 & 0.00 & 0.33 & 19.09 & 0.55\\ 
    \bottomrule
    \multirow{2}{*}{\begin{tabular}[c]{@{}l@{}}Linear Regression\\ Scaled\end{tabular}} & \multirow{2}{*}{0.00} & \multirow{2}{*}{0.00} & \multirow{2}{*}{0.03} & \multirow{2}{*}{0.00} & \multirow{2}{*}{0.00} & \multirow{2}{*}{0.00} & \multirow{2}{*}{0.00} & \multirow{2}{*}{0.06} & \multirow{2}{*}{0.47} & \multirow{2}{*}{0.08}\\ 
    & & & & & & & & & &\\
    \bottomrule
    \multicolumn{11}{l}{\textbf{Optimization Problem}}                   \\
    \bottomrule
    without L1 regularization & 0.79 & 0.38 & 0.26 & 0.05 & 0.06 & -0.24 & 0.15 & 0.30 & 2.27 & 0.36\\ 
    \bottomrule
    with L1 regularization & 0.49 & 0.27 & 0.20 & 0.02 & 0.03 & -0.16 & 0.05 & 0.22 & 1.55 & 0.39\\ 
    \bottomrule
  \end{tabular}
\end{table}

\begin{table}[htbp]
  \caption{Human values weights adjusted as strictly positive values. F1 is the average F1 score computed from performing cross validation.}
  \label{human-value-weights-positive}
  \centering
  \small
  \begin{tabular}{ l | l | l | l | l | l | l | l | l | l || l ||}
    \toprule
    \textbf{Model} & \textbf{Cr} & \textbf{Ex} & \textbf{ImRe} & \textbf{MaMe} & \textbf{Pr} & \textbf{PrRe} & \textbf{ReDi} & \textbf{SuSo} & \textbf{ReME} & \textbf{F1}\\
    \bottomrule
    \multicolumn{11}{l}{\textbf{Bradley-Terry}}                  \\
    \bottomrule
    Logistic Regression & 0.66 & 0.00 & 0.73 & 1.16 & 0.77 & 0.95 & 0.83 & 1.54 & 2.35 & 0.33\\ 
    \bottomrule
    \multirow{2}{*}{\begin{tabular}[c]{@{}l@{}}Logistic Regression\\ Cross Validation\end{tabular}} & \multirow{2}{*}{0.00} & \multirow{2}{*}{0.61} & \multirow{2}{*}{1.03} & \multirow{2}{*}{0.67} & \multirow{2}{*}{0.67} & \multirow{2}{*}{0.67} & \multirow{2}{*}{0.67} & \multirow{2}{*}{0.67} & \multirow{2}{*}{3.99} & \multirow{2}{*}{0.41}\\ 
    & & & & & & & & &\\
    \bottomrule
    \multicolumn{11}{l}{\textbf{Adapted Bradley-Terry}}                   \\
    \bottomrule
    Linear Regression & 0.00 & 0.00 & 0.10 & 0.00 & 0.00 & 0.00 & 0.00 & 0.15 & 8.74 & 0.55\\ 
    \bottomrule
    \multirow{2}{*}{\begin{tabular}[c]{@{}l@{}}Linear Regression\\ Scaled\end{tabular}} & \multirow{2}{*}{0.00} & \multirow{2}{*}{0.00} & \multirow{2}{*}{0.53} & \multirow{2}{*}{0.00} & \multirow{2}{*}{0.00} & \multirow{2}{*}{0.00} & \multirow{2}{*}{0.00} & \multirow{2}{*}{0.99} & \multirow{2}{*}{7.49} & \multirow{2}{*}{0.47} \\ 
    & & & & & & & & &\\
    \bottomrule
    \multicolumn{11}{l}{\textbf{Optimization Problem}}                   \\
    \bottomrule
    without L1 regularization & 1.50 & 0.91 & 0.73 & 0.42 & 0.43 & 0.00 & 0.57 & 0.79 & 3.65 & 0.46\\ 
    \bottomrule
    with L1 regularization & 1.42 & 0.94 & 0.80 & 0.40 & 0.42 & 0.00 & 0.46 & 0.83 & 3.73 & 0.44\\ 
    \bottomrule
  \end{tabular}
\end{table}

\begin{table}[htbp]
  \caption{Human values adjusted as positive weights and applied Laplace smoothing. F1 CV is the average F1 score computed from performing 5-fold cross validation on 130 samples.}
  \label{tab:human-value-weights-laplace-smoothing}
  \centering
  \small
  \begin{tabular}{ l | l | l | l | l | l | l | l | l | l || l ||}
    \toprule
    \textbf{Model} & \textbf{Cr} & \textbf{Ex} & \textbf{ImRe} & \textbf{MaMe} & \textbf{Pr} & \textbf{PrRe} & \textbf{ReDi} & \textbf{SuSo} & \textbf{ReME} & \textbf{F1}\\
    \bottomrule
    \multicolumn{11}{l}{\textbf{Bradley-Terry}}                  \\
    \bottomrule
    Logistic Regression & 0.66 & 0.01 & 0.73 & 1.16 & 0.77 & 0.96 & 0.84 & 1.54 & 2.34 & 0.33\\ 
    \bottomrule
    \multirow{2}{*}{\begin{tabular}[c]{@{}l@{}}Logistic Regression\\ Cross Validation\end{tabular}} & \multirow{2}{*}{0.01} & \multirow{2}{*}{0.62} & \multirow{2}{*}{1.02} & \multirow{2}{*}{0.68} & \multirow{2}{*}{0.68} & \multirow{2}{*}{0.68} & \multirow{2}{*}{0.68} & \multirow{2}{*}{0.68} & \multirow{2}{*}{3.96} & \multirow{2}{*}{0.41}\\ 
    & & & & & & & & & &\\
    \bottomrule
    \multicolumn{11}{l}{\textbf{Adapted Bradley-Terry}}                   \\
    \bottomrule
    \textbf{Linear Regression} & \textbf{0.01} & \textbf{0.01} & \textbf{0.11} & \textbf{0.01} & \textbf{0.01} & \textbf{0.01} & \textbf{0.01} & \textbf{0.16} & \textbf{8.67} & \textbf{0.57}\\ 
    \bottomrule
    \multirow{2}{*}{\begin{tabular}[c]{@{}l@{}}Linear Regression\\ Scaled\end{tabular}} & \multirow{2}{*}{0.01} & \multirow{2}{*}{0.01} & \multirow{2}{*}{0.53} & \multirow{2}{*}{0.01} & \multirow{2}{*}{0.01} & \multirow{2}{*}{0.01} & \multirow{2}{*}{0.01} & \multirow{2}{*}{0.99} & \multirow{2}{*}{7.42} & \multirow{2}{*}{0.47}\\ 
    & & & & & & & & & &\\
    \bottomrule
    \multicolumn{11}{l}{\textbf{Optimization Problem}}                   \\
    \bottomrule
    without L1 regularization & 1.50 & 0.91 & 0.73 & 0.43 & 0.44 & 0.01 & 0.57 & 0.79 & 3.63 & 0.46\\ 
    \bottomrule
    with L1 regularization & 1.41 & 0.94 & 0.80 & 0.40 & 0.42 & 0.01 & 0.47 & 0.83 & 3.70 & 0.44\\ 
    \bottomrule
  \end{tabular}
\end{table}

\section{Prompt Templates}
\label{app:prompt-templates}
\subsection{User Message prompt for Supervised Fine Tuning and Reinforcement Learning}
The following template is used as the user message prompt.

\begin{verbatim}
"""
You are a member of the scientific community tasked with peer review. 
Review the following paper content.

### Paper Content

{paper_content}
"""    
\end{verbatim}\label{user-message-prompt}

\subsection{Assistant Prompts for Supervised Fine Tuning}
The following template is used as the target response from the LLM.

\begin{verbatim}
"""
<think> {thinking traces from Sonnet 3.7} </think>

{review content}
"""    
\end{verbatim}\label{assistant-target-prompt-with-thinking-traces}

\section{Models and Dataset}
\label{app:trained-models-and-dataset}
The trained models and datasets will be publicly available at huggingface.co. Below is the list of resources:

Models
\begin{itemize}
    \item SFT on 16k reviews: pawin205/Qwen-7B-Review-ICLR-sft
    \item SFT on 90th Percentile reviews: pawin205/Qwen-7B-Review-ICLR-90th-sft
    \item GRPO Uniform (REMOR-U): pawin205/Qwen-7B-Review-ICLR-GRPO-U
    \item GRPO Human-aligned (REMOR-H): pawin205/Qwen-7B-Review-ICLR-GRPO-H
\end{itemize}
Dataset
\begin{itemize}
    \item PeerRT: ICLR 2017-2020 Reviews augmented with thinking traces, full text, and metric scores: pawin205/iclr-2017-2020-peer-review-with-thinking-trace
    \item ICLR 2017-2020 Reviews with just title and abstract with metric scores: pawin205/paper-review-pair
    \item ICLR 2017-2020 Reviews with just title, abstract, and static thinking traces with metric scores: pawin205/paper-review-pair-reason
\end{itemize}

\section{Code Repository}
\label{app:code-repository}
All scripts to process data can be found at \href{https://github.com/Khempawin/remor.git}{REPOSITORY}. This includes example scripts to load the dataset as well as loading models and perform text generation.

\section{Model performance across different metrics}
This section includes the performance of each model in tabular form.

\begin{table}[htbp]
\centering
\small
\caption{Metric-wise average reward comparison between Human, Sonnet 3.7, Deepseek and REMOR. Scores are calculated from 5.5k papers. All values had a SEM within the range of 0.001.}
\begin{tabular}{|l|c|c|c|c|c|c|c|c|c|c|c|}
\toprule
    \multirow{2}{*}{\textbf{Model}} & \multirow{2}{*}{\textbf{Cr}} & \multirow{2}{*}{\textbf{Ex}} & \multirow{2}{*}{\textbf{ImRe}} & \multirow{2}{*}{\textbf{MaMe}} & \multirow{2}{*}{\textbf{Pr}} & \multirow{2}{*}{\textbf{PrRe}} & \multirow{2}{*}{\textbf{ReDi}} & \multirow{2}{*}{\textbf{SuSo}} & \multirow{2}{*}{\textbf{ReME}} & \multirow{2}{*}{\begin{tabular}[c]{@{}l@{}}\textbf{Reward}\\ \textbf{(U)}\end{tabular}} & \multirow{2}{*}{\begin{tabular}[c]{@{}l@{}}\textbf{Reward}\\ \textbf{(H)}\end{tabular}}\\ 
    & & & & & & & & & & &\\
\bottomrule
Human & 0.212 & 0.051 & 0.118 & 0.558 & 0.129 & 0.189 & 0.193 & 0.175 & 0.026 & 1.654 & 0.285\\
\bottomrule
Sonnet 3.7 & 0.058 & 0.010 & 0.210 & 0.515 & 0.224 & 0.050 & 0.179 & 0.161 & 0.037 & 1.445 & 0.384\\
\bottomrule
DS \cite{guo2025deepseek} & 0.026 & 0.009 & 0.154 & 0.646 & 0.116 & 0.041 & 0.225 & 0.073 & 0.028 & 1.317 & 0.283\\
\bottomrule
DS SFT & 0.231 & 0.053 & 0.277 & 0.776 & 0.263 & 0.164 & 0.330 & 0.233 & 0.022 & 2.349 & 0.276\\
\bottomrule
REMOR-U & \textbf{0.310} & \textbf{0.112} & \textbf{0.564} & \textbf{0.794} & \textbf{0.605} & \textbf{0.322} & \textbf{0.594} & \textbf{0.548} & 0.034 & \textbf{3.884} & 0.470\\
\bottomrule
REMOR-H & 0.169 & 0.041 & 0.380 & 0.780 & 0.333 & 0.144 & 0.327 & 0.378 & \textbf{0.063} & 2.614 & \textbf{0.670}\\
\bottomrule
\end{tabular}%
\label{tab:metric-wise-comparison-of-training-treatments}
\end{table}

\begin{table}[htbp]
  \caption{Performance comparison with previous systems over multiple papers from ACL 2017 and NeurIPS 2019}
  \label{tab:metric-comparison-of-previous-systems}
  \centering
  \small
  \begin{tabular}{ l | l | l | l | l | l | l | l | l | l || c | c ||}
    \toprule
    \multirow{2}{*}{\textbf{Model}} & \multirow{2}{*}{\textbf{Cr}} & \multirow{2}{*}{\textbf{Ex}} & \multirow{2}{*}{\textbf{ImRe}} & \multirow{2}{*}{\textbf{MaMe}} & \multirow{2}{*}{\textbf{Pr}} & \multirow{2}{*}{\textbf{PrRe}} & \multirow{2}{*}{\textbf{ReDi}} & \multirow{2}{*}{\textbf{SuSo}} & \multirow{2}{*}{\textbf{ReME}} & \multirow{2}{*}{\begin{tabular}[c]{@{}l@{}}\textbf{Reward}\\ \textbf{(U)}\end{tabular}} & \multirow{2}{*}{\begin{tabular}[c]{@{}l@{}}\textbf{Reward}\\ \textbf{(H)}\end{tabular}}\\ 
    & & & & & & & & & & &\\
    \toprule
    \multirow{2}{*}{\begin{tabular}[c]{@{}l@{}}\textbf{Human}\\ \textbf{Reviewer}\end{tabular}} & \multirow{2}{*}{0.161} & \multirow{2}{*}{0.072} & \multirow{2}{*}{0.116} & \multirow{2}{*}{0.440} & \multirow{2}{*}{0.244} & \multirow{2}{*}{0.335} & \multirow{2}{*}{0.174} & \multirow{2}{*}{0.191} & \multirow{2}{*}{0.029} & \multirow{2}{*}{1.762} & \multirow{2}{*}{0.306}\\ 
    & & & & & & & & & & &\\
    \bottomrule
    \multicolumn{12}{l}{\textbf{Non-reasoning models}}                  \\
    \bottomrule
    Barebones & 0.047 & 0.006 & 0.209 & 0.459 & 0.307 & 0.179 & 0.213 & 0.257 & 0.022 & 1.700 & 0.269 \\
    Liang et al \cite{liang2024can} & 0.157 & 0.000 & 0.037 & 0.374 & 0.000 & 0.030 & 0.150 & 0.220 & 0.009 & 0.978 & 0.128 \\
    MARG-S \cite{d2024marg} & 0.027 & 0.019 & 0.264 & 0.500 & 0.221 & 0.139 & 0.160 & 0.234 & 0.050 & 1.615 & 0.514 \\
    MAMORX \cite{taechoyotin2024mamorx} & 0.054 & 0.024 & 0.229 & 0.527 & 0.184 & 0.201 & 0.123 & 0.250 & 0.067 & 1.658 & 0.658 \\
    \bottomrule
    \multicolumn{12}{l}{\textbf{Explicit reasoning models}}                  \\
    \bottomrule
    Sonnet 3.7 & 0.049 & 0.004 & 0.200 & 0.492 & 0.222 & 0.111 & 0.150 & 0.177 & 0.039 & 1.443 & 0.398 \\
    DeepSeek \cite{guo2025deepseek} & 0.031 & 0.017 & 0.174 & 0.688 & 0.143 & 0.181 & 0.189 & 0.138 & 0.055 & 1.616 & 0.529 \\
    DS SFT & 0.163 & 0.043 & 0.201 & 0.694 & 0.187 & 0.211 & 0.194 & 0.214 & 0.047 & 1.953 & 0.478 \\
    REMOR-U & \textbf{0.273} & \textbf{0.105} & \textbf{0.422} & \textbf{0.728} & \textbf{0.466} & \textbf{0.306} & \textbf{0.468} & \textbf{0.456} & 0.068 & \textbf{3.292} & 0.731 \\
    REMOR-H & 0.152 & 0.032 & 0.370 & 0.747 & 0.333 & 0.249 & 0.255 & 0.376 & \textbf{0.152} & 2.667 & \textbf{1.438} \\
    \bottomrule
  \end{tabular}
\end{table}

\begin{figure}[htbp]
  \centering
  \includegraphics[width=0.6\textwidth]{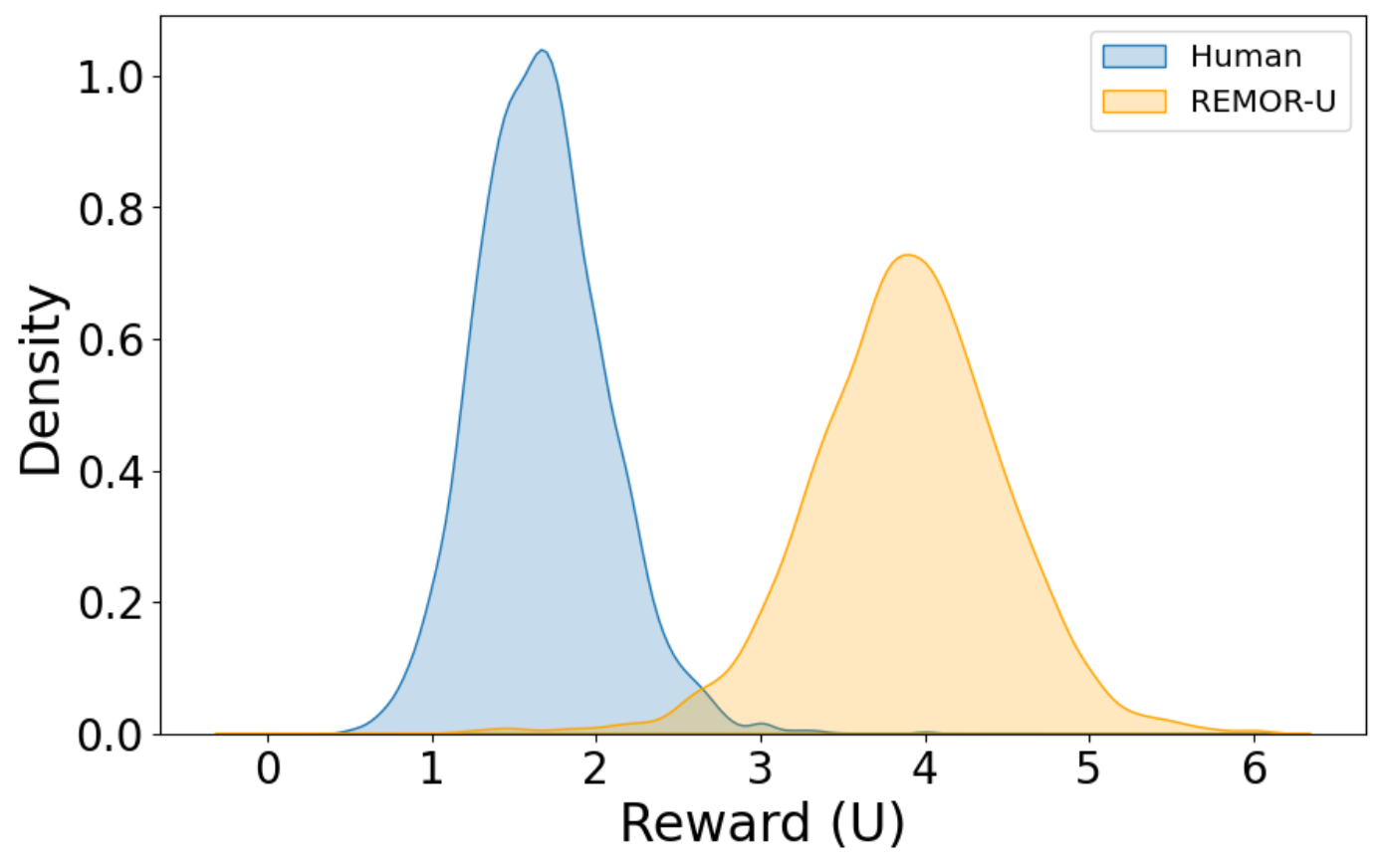}
  \caption{A distribution of Reward (U) of human and REMOR-U reviews. The plot shows that REMOR-U reviews mostly have higher uniform reward than the average human reviews. It should be noted that human reviews with a uniform reward of four exist, but in very small numbers.}
  \label{fig:reward-distribution-human-vs-remor}
\end{figure}

\section{Sample review comparison between humans and REMOR}
This section shows sample reviews from papers with a high contrast between human reviewers and REMOR. Table \ref{tab:sample_reviews_with_high_or_low_scores} contains the sampled reviews.

\begin{table}
    \centering
    \caption{Qualitative comparison of reviews with high aspect reward but low METEOR score against low aspect reward but high METEOR score}
    \tiny
        \begin{tabular}{| m{7cm} | m{7cm} |}
  \toprule
\multicolumn{2}{|c|}{\textbf{Sample Reviews from Humans and REMOR}} \\
\bottomrule
\multicolumn{2}{|c|}{↑ Human (U) ↓ REMOR-U} \\
\bottomrule
   + An interesting idea to learn the hidden state evolution and the state-observation mapping jointly + The experiments on Euler*s equation are slightly better than ResNet for 30 steps ahead forecasting in terms of MSE + \textcolor{brown}{The paper is clearly written and well-explained} - The model is not new: ResNet for state evolution and Conv-Deconv for state-observation mapping - The difference between ResNet and the proposed framework is not significant, ResNet is even better in Figure 2 - Missing an important experiment: test whether the model can generalize, that is to forecast on different initial conditions than the training dataset - How does the model compare with GANs (Y. Xie* , E. Franz* and M. Chu* and N. Thuereyy, “tempoGAN: A Temporally Coherent, Volumetric GAN for Super-resolution Fluid Flow”)?
  & 
  This paper presents a novel approach called Flow-Flow Estimation with Systematic Teacher Forcing (FST) for predicting velocity fields using both physical principles and learned controls. The authors propose FST-Accurate ODE (FST-Accurate O D ) as a practical extension of the original FST method, with improved numerical stability when applied to flows.

\textcolor{red}{...}

\textcolor{brown}{The paper is well-written, but I have some concerns about clarity}:
- Some sections feel like they could be more concise or provide clearer explanations
- The tables are dense with detailed results and would benefit from better organization
- A brief explanation of the different velocity field types (NSV, NSV, TVF) in Table 1 would help readers understand what they're measuring

The results appear promising for fluid simulations:
- Their FST model outperforms baseline approaches by 26\% at average MSE per time-step and by 83\% for velocity fields
- The approach seems particularly valuable when working with small timesteps (10 or fewer steps), but performs poorly on larger horizons.

The paper makes a valuable contribution to the field of fluid simulations, providing a new method that can be applied in various contexts. \textcolor{red}{...}

  \\
  \hline
  \multicolumn{2}{|c|}{↓ Human (U) ↑ REMOR-U} \\
\hline
GraphNVP is the first paper to introduce the concept of *invertible flow*, that is to construct the invertible mapping from latent vector z to the graph G. By constructing the mapping from G to z, GraphNVP first changes the discrete feature vector into continuous variables, then update this matrix representation by scaling and transforming functions (Eq. (2)-(5) in this GRF paper). In each iteration the matrix is only updated by one row (one slice for the tensor), while keep other rows intact. Then for constructing the inverse mapping, we can first sample a random vector and then apply the “inverse” of the update rule to recover the edge matrix and node matrix respectively. \textcolor{red}{...}

& 

\textcolor{brown}{The paper is well written but has several issues which I will now point out}: - The theoretical results in Section 3.1 are interesting, however they are quite hard to parse since there is a lack of an explanation for some terms and it*s not clear what the final result is (e.g., Theorem 1). It would be useful to provide more intuitive explanations or examples on how these can be applied - The experiments could have been better organized with clearer descriptions, as the experimental section was quite confusing - The authors should explain in greater detail why the experimental results show that GRF has lower variance than other methods (this is stated several times but explained briefly) - While I am not convinced by the results on ZINC-250k dataset which is not very challenging compared to QM9, the results are interesting and could be interesting for further research directions.
\\
\hline
\multicolumn{2}{|c|}{↑ Human (H) ↓ REMOR-H} \\
\hline

The paper considers the problem of out-of-distribution (OOD) sample detection while solving a classification task. The authors tackle the problem of OOD detection with exploiting uncertainty while passing a test sample through the neural network. They treat outputs of (some) layers in a NN as random Gaussian-distributed variables and measure uncertainty as variance of these Gaussians. Then when uncertainty is high, OOD is detected. The overall idea behind the paper could be interesting, but its realization in the current form is questionable. The paper seems totally misusing the reparameterization trick and stochastic outputs of layers in NNs. Eq. (2) is not the objective of variational inference that seems to be required for stochastic outputs and the reparameterization trick as presented before the equation. The objective misses the KL-divergence term! Without it what would stop a neural net to set sigmas to 0 and forget about the stochasticity altogether? Not to mention that the current objective is not mathematically justified. \textcolor{red}{...}

& 

The problem addressed here is interesting and has high practical importance, but the method proposed seems to be somewhat incremental, without being able to show much improvements on all datasets, and it requires more extensive experiments with different models and datasets than what have been presented so far (though some of them are provided as a part of the paper). The quality of writing is also quite poor in some parts.
\\
\hline
\multicolumn{2}{|c|}{↓ Human (H) ↑ REMOR-H} \\
\hline
Summary: The paper proposes a simple technique to address the problem introduced by Adebayo et al. that several saliency approaches do not pass sanity checks. The proposed approach computes the saliency maps for all the classes and removes the pixels that play a role in predicting several classes. Strengths: 1. Simple and intuitive approach. 2. Well written and easy to read paper. 3. The introduced approach makes Grad.Input pass the sanity checks introduced by Adebayo et al. Weaknesses: 1. For any interpretability technique, passing the sanity check is a must, but just because a saliency technique passes the sanity checks, it doesn’t mean that these maps explain the network’s decision well. 2. Lack of any quantitative evaluation (such as localization or pointing experiment) of their approach. 3. Failure to show if the resultant maps are class-discriminative. Show performance on images with multiple classes. 4. \textcolor{red}{...}
& 
This paper discusses a relatively simple modification to existing methods that helps saliency maps pass sanity checks proposed by Adebayo et al., 2018. The approach is interesting, and the idea of using competition among labels seems to make sense in order to remove features which are deemed irrelevant or coincidental by humans. It is well written paper discussing an interesting modification and its effect on existing methods (but not really a novel method), which I think should be published after more analysis and experimentation would have been possible, such as comparison with other existing approaches for saliency evaluation. However the modifications introduced here seem to work well in practice. The experimental results are promising although some of them could be better explained by the theory presented earlier in the paper (section 4). Overall I think this is a solid submission that addresses an interesting topic, though I feel it would be more interesting with a more theoretical justification for why the method works and comparison to other existing approaches for saliency evaluation. \textcolor{red}{...}
\\
\hline
\end{tabular}
    
    \label{tab:sample_reviews_with_high_or_low_scores}
\end{table}

\newpage
\thispagestyle{empty}
\newpage

\end{document}